\documentclass{article}
\usepackage[utf8]{inputenc}
\usepackage[T1]{fontenc}
\usepackage{times}
\usepackage{helvet}
\usepackage{courier}

\usepackage{amsmath}
\usepackage{amssymb}
\usepackage{amsfonts}

\usepackage{booktabs}
\usepackage{multirow}
\usepackage{makecell}
\usepackage{array}
\usepackage{tabularx}

\usepackage{graphicx}
\usepackage{xcolor}
\usepackage{caption}

\usepackage{algpseudocode} 
\usepackage{algorithm}
\usepackage{enumitem}
\usepackage{tcolorbox}
\usepackage{float}
\usepackage{nicefrac}
\usepackage{microtype}
\usepackage{mathtools}

\usepackage{url}
\usepackage[colorlinks=true, linkcolor=blue, citecolor=blue, urlcolor=blue]{hyperref}

\usepackage{tikz}

\usepackage[numbers,square]{natbib}

\usepackage{arxiv}

\usepackage{amsthm}

\title{NeuroSymb-MRG: Differentiable Abductive Reasoning with Active Uncertainty Minimization for Radiology Report Generation}

\author{
    Rong Fu\thanks{Corresponding author: mc46603@um.edu.mo} \\
    University of Macau \\
    \texttt{mc46603@um.edu.mo} \\
    \And
    Yiqing Lyu \\
    Tsinghua University \\
    \texttt{lvyq22@mails.tsinghua.edu.cn} \\
    \And
    Chunlei Meng \\
    Fudan University \\
    \texttt{clmeng23@m.fudan.edu.cn} \\
    \And
    Muge Qi \\
    Peking University \\
    \texttt{2301210659@stu.pku.edu.cn} \\
    \And
    Yabin Jin \\
    The First People's Hospital of Foshan \\
    \texttt{jinyabin1990@qq.com} \\
    \And
    Qi Zhao \\
    University of Macau \\
    \texttt{qizhao@um.edu.mo} \\
    \And
    Li Bao \\
    Capital Medical University \\
    \texttt{baolilq909@sina.com} \\
    \And
    Juntao Gao \\
    Tsinghua University \\
    \texttt{jtgao@mail.tsinghua.edu.cn} \\
    \And
    Fuqian Shi \\
    Rutgers Cancer Institute, NJ, USA; and NYU Langone Health, NY, USA \\
    \texttt{fuqian.shi@med.nyu.edu} \\
    \And
    Nilanjan Dey \\
    Techno International New Town, Kolkata, India \\
    \texttt{nilanjan.dey@tint.edu.in} \\
    \And
    Wei Luo \\
    The First People's Hospital of Foshan \\
    \texttt{luowei\_421@163.com} \\
    \And
    Simon Fong \\
    University of Macau \\
    \texttt{ccfong@um.edu.mo}
}

\renewcommand{\headeright}{}
\renewcommand{\undertitle}{}
\renewcommand{\shorttitle}{NeuroSymb-MRG}

\hypersetup{
    pdftitle={NeuroSymb-MRG: Differentiable Abductive Reasoning with Active Uncertainty Minimization for Radiology Report Generation},
    pdfsubject={Computer Science - Computer Vision and Pattern Recognition (cs.CV)}, 
    pdfauthor={Rong Fu, Yiqing Lyu, Chunlei Meng, Muge Qi, Yabin Jin, Qi Zhao, Li Bao, Juntao Gao, Fuqian Shi, Nilanjan Dey, Wei Luo, Simon Fong},
    pdfkeywords={NeuroSymbolic radiology report generation, differentiable abductive reasoning, active uncertainty minimization, retrieval-augmented generation, medical concept extraction},
    pdfstartview={FitH},
    colorlinks=true,
    linkcolor=red,
    citecolor=green,
    filecolor=magenta,
    urlcolor=cyan
}

\begin{document}
\maketitle

\begin{abstract}
Automatic generation of radiology reports seeks to reduce clinician workload while improving documentation consistency. Existing methods that adopt encoder-decoder or retrieval-augmented pipelines achieve progress in fluency but remain vulnerable to visual-linguistic biases, factual inconsistency, and lack of explicit multi-hop clinical reasoning. We present \textsc{NeuroSymb-MRG}, a unified framework that integrates NeuroSymbolic abductive reasoning with active uncertainty minimization to produce structured, clinically grounded reports. The system maps image features to probabilistic clinical concepts, composes differentiable logic-based reasoning chains, decodes those chains into templated clauses, and refines the textual output via retrieval and constrained language-model editing. An active sampling loop driven by rule-level uncertainty and diversity guides clinician-in-the-loop adjudication and promptbook refinement. Experiments on standard benchmarks demonstrate consistent improvements in factual consistency and standard language metrics compared to representative baselines.
\end{abstract}


\section{Introduction}
Medical report writing is a central component of clinical workflows and the primary vehicle for recording imaging findings, diagnostic impressions, and follow-up recommendations. Producing high-quality radiology reports is time-consuming and scales poorly with clinical demand, which motivates automated radiology report generation research that can reduce clinician burden while maintaining clinical accuracy.

Prior work adapted general image captioning and sequence generation paradigms to this domain. Early encoder-decoder models and attention mechanisms established the baseline for image-to-text translation \cite{vinyals2015show,vaswani2017attention}. Subsequent improvements targeted the particularities of radiology by introducing adaptive attention, bottom-up visual features, memory and meshed-memory modules, and retrieval-augmented or prototype-driven cross-modal fusion \cite{lu2017knowing,anderson2018bottom,cornia2020meshed,chen2020generating,chen2021cross,wang2022cross}. More recent approaches explore frozen large language models augmented with lightweight visual alignment modules and prompt tuning to leverage rich pre-trained linguistic priors \cite{wang2023r2gengpt,li2025multimodal}. These advances improve fluency and enable stronger text generation when image-text data are limited, yet important gaps remain. Publicly used datasets such as MIMIC-CXR and IU X-ray provide essential scale but also expose distributional imbalance and label noise that hinder reliable end-to-end learning \cite{johnson2019mimic,demner2016preparing}. 

Modern medical report generation methods face three recurring limitations. First, text decoders trained from limited paired data can produce fluent but factually incorrect statements and hallucinations. Second, purely statistical encoders lack explicit multi-hop reasoning and thus struggle to produce interpretable, clause-level justifications for diagnostic claims. Third, dataset imbalance and clinical rare-event sparsity make it hard to guarantee robust coverage for low-prevalence conditions. Prior remedies include cross-modal causal interventions, memory networks, and diagnosis-driven prompts, but these techniques do not provide an integrated mechanism to produce human-interpretable reasoning chains while quantifying uncertainty at the rule level \cite{liu2021exploring,chen2023cross,jin2024promptmrg,hirsch2024medrat}. 

These observations motivate a different design point: combine explicit differentiable neuro-symbolic reasoning, which yields structured and interpretable intermediate representations, with retrieval-augmented generation and an uncertainty-aware active sampling loop that prioritizes clinician review for high-value cases. Differentiable logical operators permit end-to-end optimization while preserving the ability to decode hard rule instantiations for explanation. Active uncertainty minimization at the reasoning level improves both data efficiency and factual reliability by directing human effort to ambiguous or high-impact examples, which is critical given the clinical cost of errors \cite{liang2025ai,agishev2025fusionforce,wasilewski2025deep,junhwan2022uncertainty,liang2025uncertainty}. 

Our contributions are as follows. First, we formulate radiology report generation as a structured mapping from image to a triple of probabilistic concepts, a differentiable abductive reasoning chain, and a templated textual draft, and we implement a differentiable logic layer that composes soft logical operators into multi-hop rules suitable for medical claims. Second, we design a hybrid generation pipeline that decodes soft rule activations into canonical clause templates, augments these drafts with retrieval evidence and constrained LLM paraphrasing, and employs a verifier to detect high-risk contradictions. Third, we introduce an active uncertainty minimization mechanism that measures rule-level predictive entropy with Monte Carlo dropout, selects diverse high-entropy cases using a k-center strategy, and uses a learned feedback simulator to accelerate clinician-in-the-loop refinement. Finally, we validate the design on standard benchmarks and show that integrating explicit reasoning and rule-level active sampling improves factual consistency and common language metrics relative to strong baselines. The experimental setup, ablations, and implementation details are informed by common practices and datasets in the field. 


\section{Related Work}

\subsection{Radiology Report Generation Strategies}
The automated construction of clinical narratives from diagnostic imagery has evolved significantly. Early attempts adapted general image captioning frameworks, such as the encoder-decoder paradigm \cite{vinyals2015show} and attention-driven models \cite{vaswani2017attention}, to the medical domain. To address the unique spatial requirements of radiology, researchers introduced adaptive sentinel mechanisms \cite{lu2017knowing} and bottom-up attention \cite{anderson2018bottom}. Recent breakthroughs have leveraged the Transformer architecture, with innovations including meshed-memory modules \cite{cornia2020meshed} and memory-driven relational modeling \cite{chen2020generating} to maintain clinical consistency. Specifically, cross-modal memory networks \cite{chen2021cross} and hierarchical alignment strategies \cite{you2021aligntransformer} have been utilized to bridge the semantic gap between visual findings and anatomical descriptions. To further enhance diagnostic accuracy, progressive generation techniques \cite{nooralahzadeh2021progressive} and reinforced cross-modal alignment \cite{qin2022reinforced} have been proposed. Furthermore, the integration of Large Language Models (LLMs) via customized prompt tuning \cite{li2025multimodal} and frozen weight architectures \cite{wang2023r2gengpt} marks a shift toward utilizing massive pre-trained linguistic knowledge for medical report synthesis.

\subsection{Neural-Symbolic AI and Differentiable Reasoning}
The integration of connectionist learning with symbolic logic represents a pivotal shift in AI reasoning \cite{liang2025ai}. Traditional neural networks often struggle with the transparency required for clinical decision-making. Neural-symbolic systems aim to reconcile these paradigms by embedding logical constraints within differentiable frameworks. Recent research has explored the use of fuzzy Zadeh’s T-norms to develop deep differentiable logic gate networks \cite{wasilewski2025deep}, which allow for gradient-based optimization of Boolean operations. In structured domains, end-to-end differentiable neural-symbolic layers have been successfully applied to tasks like trajectory prediction \cite{agishev2025fusionforce}. For higher-level cognitive tasks, balanced neuro-symbolic approaches have been developed to handle commonsense abductive logic \cite{cotnareanu2026balanced}, providing a roadmap for interpreting complex medical observations through logical abduction. Unlike pure black-box models, these architectures prioritize the preservation of logical rules throughout the learning process.

\subsection{Uncertainty Estimation and Active Learning}
Reliability in medical AI is inextricably linked to the model's ability to quantify its own confidence. Uncertainty estimation using Bayesian dropout has proven effective in high-stakes fields like seismic inversion \cite{junhwan2022uncertainty} and is increasingly applied to medical diagnostics. By identifying regions of high predictive variance, models can mitigate the risks of hallucination. Active learning frameworks further exploit these uncertainty metrics to optimize data selection; for instance, combining diversity-based sampling with uncertainty estimates has shown significant promise in tracking applications \cite{liang2025uncertainty}. Within the healthcare sector, multi-agent reinforcement learning \cite{sheikh2025advancing} and retrieval-augmented structuring \cite{jiang2025retrieval} have been proposed to manage complex system-of-systems environments. Our work extends these principles by applying active uncertainty minimization specifically to the reasoning chains within the neuro-symbolic module.

\subsection{Pre-training and Knowledge Integration}
The availability of large-scale datasets such as MIMIC-CXR \cite{johnson2019mimic} and curated radiology examinations \cite{demner2016preparing} has laid the foundation for modern medical vision systems. Self-supervised learning (SSL) has become a standard for medical image analysis \cite{rani2024self}, enabling the extraction of robust pathological features without exhaustive manual labeling. Beyond feature extraction, distilling posterior and prior knowledge \cite{liu2021exploring} and employing cross-modal causal interventions \cite{chen2023cross} have been explored to ensure that generated reports are grounded in evidence rather than spurious correlations. Recent advances also emphasize the importance of medical semantic assistance \cite{wang2022medical} and prototype-driven networks \cite{wang2022cross} to standardize the reporting process across diverse clinical settings. Furthermore, semi-supervised approaches using graph-guided consistency \cite{zhang2023semi} and diagnosis-driven prompts \cite{jin2024promptmrg} continue to refine the boundary between data-driven learning and expert-informed reasoning.

\begin{figure*}[t]
  \centering
  \includegraphics[width=0.95\textwidth]{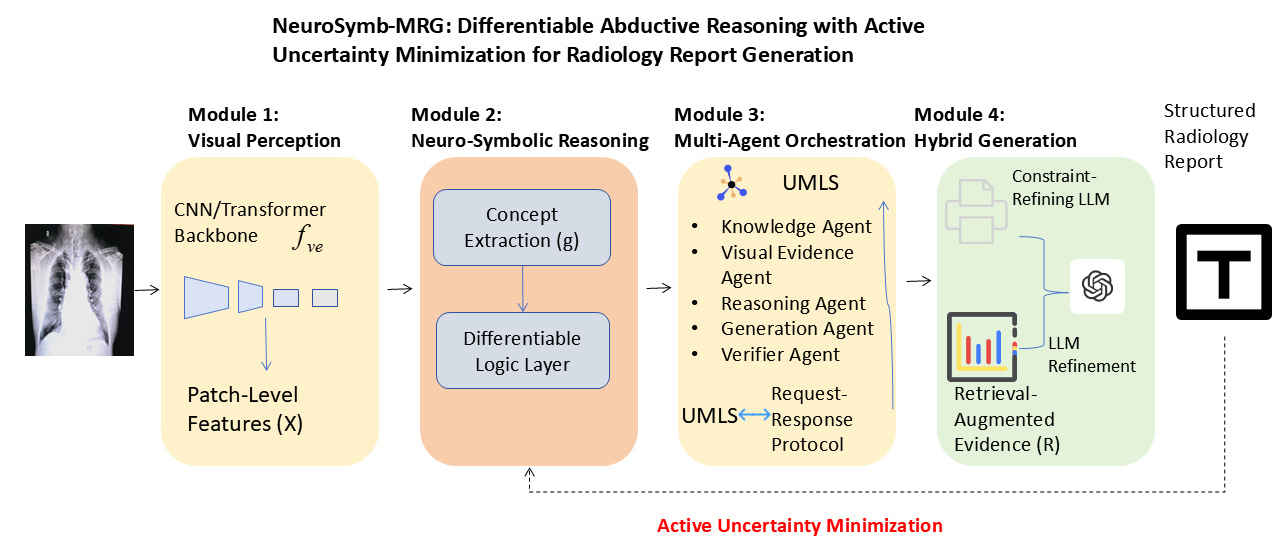} 
  \caption{Architectural overview of the \textbf{\textsc{NeuroSymb-MRG}} framework for transparent and clinically grounded radiology report generation. 
  The pipeline initiates with \textbf{Visual Perception}, utilizing a self-supervised visual encoder $f_{\mathrm{ve}}$ to extract patch-level features $X$. 
  In the \textbf{Neuro-Symbolic Reasoning} module, these features are mapped to probabilistic concept activations $\hat{c}$, which serve as leaves for a \textbf{Differentiable Logic Layer} that composes multi-hop reasoning chains through soft logical operators (AND, OR, NOT). 
  The \textbf{Multi-Agent Orchestration} hub employs a request-response protocol among specialized agents (Knowledge, Verifier, Reasoning) to cross-reference findings against a \textbf{Medical Knowledge Graph (UMLS)}. 
  The \textbf{Hybrid Generation} phase merges rule-decoded clauses $\widehat{\mathcal{L}}$ with retrieval-augmented evidence $\mathcal{R}$ via a template inventory $\mathcal{T}$ and constrained LLM refinement. 
  Finally, an \textbf{Active Uncertainty Minimization} loop utilizes predictive entropy and a \textbf{Feedback Simulator} to identify high-value cases for clinician adjudication, iteratively refining the promptbook and model parameters.} 
  \label{fig:neurosymb_mrg_framework}
\end{figure*}
\section{Methodology}

This section formalizes the radiology report generation problem and details the proposed \textsc{NeuroSymb-MRG} architecture.

\subsection{Problem statement}
We consider a single radiological study represented by image \(I \in \mathbb{R}^{H\times W\times C}\) and its associated ground-truth structured report \(R^\star\). The structured report is modeled as a tuple containing extracted concepts, an explicit reasoning chain, and a templated textual impression. The learning objective is to construct a mapping
\begin{align}
\mathcal{M}: I \mapsto \widehat{R} = \big(\widehat{C}, \widehat{\mathcal{L}}, \widehat{T}\big),
\label{eq:mapping}
\end{align}
where \(\widehat{C}\) denotes the set of probabilistic concept activations, \(\widehat{\mathcal{L}}\) denotes a multi-hop neuro-symbolic reasoning chain, and \(\widehat{T}\) denotes the final structured textual output produced by template filling and optional LLM refinement. Here \(I\) is the input image and \(\widehat{R}\) is the generated structured report. 

\subsection{System overview}
The proposed system decomposes into five tightly coupled modules: a visual representation module that produces domain-robust embeddings; a concept extraction module that maps visual embeddings to clinically meaningful concept probabilities; a differentiable logic module that composes concepts into soft logical rules and multi-hop chains; a retrieval-augmented template filling and LLM refinement module that generates the textual impression conditioned on the reasoning chain; and a multi-agent coordinator that manages communication, verification, and active sampling. The subsequent subsections specify architecture, operators, hyperparameters, and training objectives.

\subsection{Visual encoder and projection}
We employ a self-supervised pretrained visual encoder \(f_{\mathrm{ve}}\) to produce patch-level features:
\begin{align}
X &= f_{\mathrm{ve}}(I) \in \mathbb{R}^{M\times C'},
\label{eq:vision}
\end{align}
where \(M\) is the number of visual patches and \(C'\) is the patch feature dimension. The encoder is pretrained with pathology-preserving augmentations using contrastive/self-distillation objectives. A linear projection \(f_{\mathrm{proj}}\) maps patch features to the LLM token embedding space:
\begin{align}
V &= f_{\mathrm{proj}}(X) \in \mathbb{R}^{M\times D},
\label{eq:proj}
\end{align}
where \(D\) is the dimensionality of the LLM token embeddings and \(V\) are the visual tokens used for retrieval and conditioning. 

\subsection{Concept extraction architecture}
The concept predictor \(g\) is implemented as a transformer-based head followed by a compact MLP classifier. Specifically, the patch features \(X\) are input to a single transformer encoder block with \(N_h=8\) attention heads producing attended patch features \(X' \in \mathbb{R}^{M\times C'}\):
\begin{align}
X' &= \mathrm{TransEnc}(X; N_h=8).
\label{eq:transenc}
\end{align}
Here \(\mathrm{TransEnc}\) denotes a standard multi-head self-attention encoder block with layer normalization and a residual feed-forward sublayer. A pooled vector \(\bar{x}\in\mathbb{R}^{C'}\) is computed by global average pooling over the spatial dimension:
\begin{align}
\bar{x} &= \tfrac{1}{M}\sum_{m=1}^{M} X'_{m}.
\label{eq:pool}
\end{align}
The pooled representation is passed through a two-layer MLP with hidden dimension \(512\) and sigmoid outputs to obtain \(K\) probabilistic concept activations:
\begin{align}
\hat{c} &= \sigma\big(W_2\; \mathrm{ReLU}(W_1 \bar{x} + b_1) + b_2\big) \in [0,1]^K,
\label{eq:concept}
\end{align}
where \(K\) is the number of clinical concepts, \(W_1,W_2,b_1,b_2\) are learnable parameters, and \(\sigma\) is the element-wise sigmoid. To handle positive class sparsity we use focal loss with parameter \(\gamma=2\) for \(\mathcal{L}_{\mathrm{concept}}\). 

\subsection{Differentiable logic layer: operators and composition}
The differentiable logic module \(r_\theta\) is specified as a soft, parameterized operator network that implements a small algebra of continuous logical operators and composes rules as soft trees. Conjunction uses the product t-norm:
\begin{align}
\mathrm{AND}(a,b) &= a \cdot b,
\label{eq:and}
\end{align}
where \(a,b\in[0,1]\). Disjunction is implemented as probabilistic sum:
\begin{align}
\mathrm{OR}(a,b) &= a + b - a b,
\label{eq:or}
\end{align}
and negation is the standard complement:
\begin{align}
\mathrm{NOT}(a) &= 1 - a.
\label{eq:not}
\end{align}
Each logical rule is represented as a soft tree whose leaves are concept activations \(\hat{c}_k\) and whose internal nodes are operator units with learnable scalar gating weights \(w\in[0,1]\). The rule activation vector is computed as
\begin{align}
\bar{r}_j &= r_{\theta,j}(\hat{c}) = \Phi_j\big(\hat{c}; \theta_j\big),
\label{eq:ruleact}
\end{align}
where \(\Phi_j\) denotes the parameterized composition corresponding to rule \(j\) and \(\theta_j\) collects its internal weights. In the implementation, each internal operator node computes a weighted operator of its two inputs, for example:
\begin{align}
\mathrm{Node}(u,v;\alpha) &= \alpha\cdot \mathrm{AND}(u,v) + (1-\alpha)\cdot \mathrm{OR}(u,v),
\label{eq:node}
\end{align}
where \(\alpha\in[0,1]\) is learnable and blends conjunction and disjunction behaviors. For interpretability we maintain discrete proxies \(q(\hat{c})\in\{0,1\}^K\) obtained via thresholding; during backward propagation we use a straight-through estimator that routes gradients through continuous activations while preserving a hard selection for decode-time rule instantiation. 

The choice of product t-norm and probabilistic sum is motivated by numerical stability and probabilistic interpretability in the medical imaging context; product t-norm yields multiplicative evidence accumulation for co-occurring findings while probabilistic sum preserves independence assumptions for alternatives.

\subsection{Rule decoder and template inventory}
The rule decoder \(d\) maps activated soft rules \(\bar{r}\) to a canonical sequence of human-readable clauses. The decoder is implemented as a hybrid mechanism combining a learned classifier that selects a template identifier and a deterministic slot-filling module. The system maintains a templated inventory \(\mathcal{T}\) of \(T\) canonical clause skeletons, where \(T\) is chosen to cover common diagnostic patterns; in practice we provision \(T\approx 120\) templates that include slots for laterality, region, measurement, and uncertainty qualifiers. Given the top \(H\) rules by activation, the decoder computes
\begin{align}
\widehat{\mathcal{L}} &= d(\bar{r}) = \big(l_1, l_2, \ldots, l_H\big), \qquad l_i = \mathrm{Fill}\big(\tau_{p(i)}, \mathrm{Slots}(\hat{c}, s_i)\big),
\label{eq:decode}
\end{align}
where \(p(i)\) indexes the selected template for clause \(i\), \(\tau\) is a template skeleton, and \(\mathrm{Slots}(\cdot)\) extracts slot values (for example, laterality or size) from \(\hat{c}\) and small numeric regressors. When slot conflicts arise between multiple activated rules, the decoder resolves them by a weighted voting mechanism where votes are weighted by rule activation magnitudes \(\bar{r}_j\) and by knowledge-graph consistency scores provided by the knowledge agent (see Section on multi-agent orchestration). In cases where conflicts cannot be resolved deterministically, clauses are emitted with an explicit uncertainty qualifier (for example, ``possible'', ``cannot exclude'').

\subsection{Retrieval-augmented template filling and LLM refinement}
A vector database stores paired image-level embeddings and high-quality report fragments along with associated soft-prompts. At inference, we compute a pooled visual embedding \(v=\mathrm{pool}(V)\) and retrieve the top-\(N_r\) exemplars:
\begin{align}
\mathcal{R} &= \mathrm{Retrieve}(v; N_r),
\label{eq:retrieve}
\end{align}
where \(N_r\) is the retrieval size and \(\mathcal{R}\) are the retrieved report snippets. Retrieved fragments are treated as candidate evidence rather than verbatim insertions. The template filler \(\mathcal{F}\) produces a structured draft by combining the decoded clauses \(\widehat{\mathcal{L}}\), the retrieved evidence \(\mathcal{R}\), and the selected templates \(\mathcal{T}\):
\begin{align}
\widehat{T}_{\mathrm{draft}} &= \mathcal{F}\big(\widehat{\mathcal{L}}, \mathcal{R}, \mathcal{T}\big).
\label{eq:fill}
\end{align}
The filler marks provenance for each slot (rule-derived or retrieval-derived) and attaches a confidence score. To reduce copy-paste contradictions, an LLM-based constrained paraphrasing step is optionally applied to retrieved fragments so that the final text uses consistent phrasing while preserving factual content. A verifier agent (Section on orchestration) scores entailment between retrieval fragments and rule-derived claims; if contradictions exceed a threshold the slot is flagged for clinician review rather than auto-filled.

\subsection{Active sampling, uncertainty quantification, and feedback simulator}
Active sampling follows a two-stage policy. Uncertainty is measured by predictive entropy over rule activations, estimated via \(T_{\mathrm{MC}}=5\) stochastic forward passes with dropout enabled:
\begin{align}
H[\bar{r}] &= -\sum_{j=1}^{R} \bar{r}_j \log \bar{r}_j, \qquad \bar{r} \approx \tfrac{1}{T_{\mathrm{MC}}}\sum_{t=1}^{T_{\mathrm{MC}}} \bar{r}^{(t)}.
\label{eq:entropy}
\end{align}
Here \(\bar{r}^{(t)}\) denotes the \(t\)-th dropout sample and \(R\) is the number of rules. Diversity is promoted by solving a greedy k-center approximation in the joint embedding space formed by concatenating pooled visual embedding and concept vector, i.e. embeddings \([v; \hat{c}]\). In each active round the system selects \(k=16\) samples by first ranking by entropy and then applying the greedy k-center procedure to the top \(M_{\mathrm{cand}}\) candidates, where \(M_{\mathrm{cand}}\) is an operating parameter that trades off annotation budget and selection quality.

To accelerate human-in-the-loop iterations we train a feedback simulator that proposes clinician-like edits. The simulator is implemented as a T5-small sequence-to-sequence model fine-tuned on historical (draft, correction) pairs extracted from MIMIC-CXR. Fine-tuning uses AdamW with learning rate \(5\times 10^{-5}\), batch size 32, and 5 epochs. The simulator is used to pre-filter low-value queries and to propose refined drafts; clinician approval remains required for any changes to the promptbook or template library.

\subsection{Multi-agent orchestration and communication protocol}
The system operationalizes five agents: a visual evidence agent, a knowledge agent, a reasoning agent, a generation agent, and a verifier agent. Agents exchange structured messages in a small JSON-like schema containing fields \textit{evidence}, \textit{claims}, \textit{templates}, \textit{confidence}, and \textit{action\_req}. Communication follows a request-response protocol where each message is timestamped and signed with agent identity. Conflict resolution is implemented by a coordinator that aggregates candidate claims using weighted voting where weights are computed from agent confidences and domain priorities; mathematically, for a slot \(s\) with candidate values \(\{x_i\}\) and agent confidences \(\{\gamma_i\}\), the coordinator selects
\begin{align}
x^\star_s &= \arg\max_{x} \sum_{i: x_i = x} \gamma_i,
\label{eq:coord}
\end{align}
where \(\gamma_i\in[0,1]\) is the confidence reported by agent \(i\). The knowledge agent consults a cached subset of UMLS to validate co-occurrence plausibility and to provide contradiction penalties that reduce the effective \(\gamma_i\) for conflicting claims. The verifier implements an entailment model to detect factual contradictions and returns a binary review flag if the contradiction score exceeds a pre-set threshold.

\subsection{Training objectives and dynamic loss weighting}
Model parameters are optimized under a composite loss:
\begin{align}
\mathcal{L} &= \lambda_{\mathrm{rep}}\mathcal{L}_{\mathrm{SSL}} + \lambda_{c}\mathcal{L}_{\mathrm{concept}} + \lambda_{t}\mathcal{L}_{\mathrm{task}} + \lambda_{r}\mathcal{L}_{\mathrm{rule}} + \lambda_{\theta}\|\theta\|^2,
\label{eq:loss}
\end{align}
where each term corresponds to representation pretraining, concept supervision, token/template reconstruction, rule fidelity, and parameter regularization, respectively. Each \(\lambda\) is initialized to balance gradient magnitudes; specifically we initialize \(\lambda_{\mathrm{rep}}=1.0\), \(\lambda_{c}=2.0\), \(\lambda_{t}=1.0\), \(\lambda_{r}=1.0\), and \(\lambda_{\theta}=10^{-4}\). To avoid domination by a single objective we adopt a homoscedastic uncertainty weighting scheme where task weights are learned as \(\lambda_i = 1/(2\sigma_i^2)\) and the training objective includes \(\log\sigma_i\) regularizers, i.e.
\begin{align}
\tilde{\mathcal{L}} &= \sum_{i\in\{\mathrm{rep},c,t,r\}} \frac{1}{2\sigma_i^2}\mathcal{L}_i + \log\sigma_i + \lambda_{\theta}\|\theta\|^2,
\label{eq:uncertainty_weight}
\end{align}
where each \(\sigma_i>0\) is a learnable scalar that adapts the relative importance of losses during training. The \(\sigma_i\) are initialized so that \(1/(2\sigma_i^2)\) matches the initial \(\lambda_i\) settings. Optionally, one can employ GradNorm as an alternative to actively balance gradient magnitudes across tasks.

\subsection{Algorithm (training and inference skeleton)}
The following algorithm references the main modules and equations above and presents the training/inference control flow.

\begin{algorithm}[H]
\caption{Training and inference skeleton for \textsc{NeuroSymb-MRG}}
\begin{algorithmic}[1]
\State \textbf{Input:} unlabeled set $\mathcal{D}_u$, labeled set $\mathcal{D}_l$ (optional), promptbook $P$, template library $\mathcal{T}$
\For{each training iteration}
    \State Sample mini-batch $I \sim \mathcal{D}_u$ and compute $X=f_{\mathrm{ve}}(I)$ using \eqref{eq:vision}
    \State Update visual encoder with SSL loss $\mathcal{L}_{\mathrm{SSL}}$ using \eqref{eq:loss}
    \If{$\mathcal{D}_l$ contains labeled examples in the batch}
        \State Compute attended features $X'=\mathrm{TransEnc}(X)$ and pooled $\bar{x}$ via \eqref{eq:transenc}--\eqref{eq:pool}
        \State Compute concepts $\hat{c}$ via \eqref{eq:concept}
        \State Compute soft rule activations $\bar{r}=r_\theta(\hat{c})$ via \eqref{eq:ruleact}
        \State Decode reasoning chain $\widehat{\mathcal{L}}=d(\bar{r})$ via \eqref{eq:decode}
        \State Retrieve exemplars $\mathcal{R}=\mathrm{Retrieve}(v;N_r)$ per \eqref{eq:retrieve}
        \State Fill templates $\widehat{T}=\mathcal{F}(\widehat{\mathcal{L}},\mathcal{R},\mathcal{T})$ per \eqref{eq:fill}
        \State Compute supervised losses $\mathcal{L}_{\mathrm{concept}}, \mathcal{L}_{\mathrm{rule}}, \mathcal{L}_{\mathrm{task}}$
        \State Optimize composite objective \(\tilde{\mathcal{L}}\) in \eqref{eq:uncertainty_weight}
    \EndIf
    \State Periodically update promptbook $P$ and vector DB with clinician-approved exemplars
    \State Active sampling: compute entropies via \eqref{eq:entropy} and select $k=16$ samples per round with greedy k-center
    \State Present structured drafts to clinicians and store (draft, correction) for feedback simulator fine-tuning
\EndFor
\State \textbf{Inference:} For a test image $I$, compute $X$ and $V$, extract $\hat{c}$ and $\bar{r}$, decode $\widehat{\mathcal{L}}$, retrieve $\mathcal{R}$, fill templates $\widehat{T}$, optionally apply constrained LLM refinement, and run verifier checks; if verifier raises a review flag, present for clinician adjudication.
\end{algorithmic}
\end{algorithm}

\subsection{Implementation notes and hyperparameters}
Default hyperparameters recommended for reproducible experiments are: transformer encoder with one block and eight heads, MLP hidden dim 512, focal loss \(\gamma=2\) for concept supervision, MC-dropout passes \(T_{\mathrm{MC}}=5\) for entropy estimation, retrieval size \(N_r=5\), active selection size \(k=16\) per round, and initial loss weights as given in \eqref{eq:loss}. The feedback simulator uses T5-small with AdamW learning rate \(5\times 10^{-5}\), batch size 32, and 5 finetuning epochs. The coordinator uses cached UMLS lookups with a local TTL-cached index to bound latency.

\section{Experiments}

\subsection{Datasets and evaluation metrics}
We evaluate \textsc{NeuroSymb-MRG} on two widely used radiology report benchmarks. The MIMIC-CXR\cite{johnson2019mimic} corpus contains 377,110 chest X-ray images paired with 227,835 radiology reports. Following standard splits, the experiments use 368,960 images and 222,758 reports for training, 2,991 images and 1,808 reports for validation, and 5,159 images and 3,269 reports for testing. The IU X-ray\cite{demner2016preparing} dataset comprises 7,470 images with 3,955 associated reports and we employ a patient-disjoint 7:1:2 split for training, validation, and testing. For quantitative evaluation we report BLEU-1 through BLEU-4, ROUGE-L and METEOR.

\subsection{Implementation details}
The visual backbone is ConvNeXt-Tiny pretrained on ImageNet-1K. Feature maps of size $7\times 7$ with 768 channels are extracted and projected into the LLM token space; in practice $M=49$, $C'=768$ and $D=4096$. The promptbook contains 50 learnable prompts. All models are trained with the Adam optimizer using a batch size of 6 and an initial learning rate of $1\times10^{-4}$. Training runs for 15 epochs on IU X-ray and 10 epochs on MIMIC-CXR. Experiments are implemented in PyTorch and executed on four NVIDIA GeForce RTX 4090 GPUs. Active sampling uses five MC-dropout passes for uncertainty estimation and a per-round selection budget of 16 examples.

\subsection{Main quantitative results}
Table~\ref{tab:main_results} summarizes the main numerical comparison between \textsc{NeuroSymb-MRG} and representative prior methods on IU X-ray and MIMIC-CXR. The table reports BLEU-1 through BLEU-4, ROUGE-L and METEOR for each method. Best and second-best values are indicated in bold and underlined text respectively. The reproduced baseline numbers and additional entries are shown for completeness. Across both datasets and all metrics, \textsc{NeuroSymb-MRG} yields consistent and substantial improvements.

\begin{table}[!ht]
\centering
\caption{Quantitative comparison with representative prior methods on IU X-ray and MIMIC-CXR. Best and second-best values are in \textbf{bold} and \underline{underlined}, respectively.}
\label{tab:main_results}
\resizebox{\textwidth}{!}{%
\begin{tabular}{l|cccccc|cccccc}
\toprule
\multirow{2}{*}{Method} & \multicolumn{6}{c|}{IU X-ray\cite{demner2016preparing}} & \multicolumn{6}{c}{MIMIC-CXR\cite{johnson2019mimic}} \\
\cmidrule(lr){2-7} \cmidrule(lr){8-13}
 & B-1 & B-2 & B-3 & B-4 & R-L & MTR & B-1 & B-2 & B-3 & B-4 & R-L & MTR \\
\midrule
Show-Tell\cite{vinyals2015show} & 0.243 & 0.130 & 0.108 & 0.078 & 0.307 & 0.157 & 0.308 & 0.190 & 0.125 & 0.088 & 0.256 & 0.122 \\
Transformer\cite{vaswani2017attention} & 0.372 & 0.251 & 0.147 & 0.136 & 0.317 & 0.168 & 0.316 & 0.199 & 0.140 & 0.092 & 0.267 & 0.129 \\
Att2in\cite{rennie2017self} & 0.248 & 0.134 & 0.116 & 0.091 & 0.309 & 0.162 & 0.314 & 0.198 & 0.133 & 0.095 & 0.264 & 0.122 \\
AdaAtt\cite{lu2017knowing} & 0.284 & 0.207 & 0.150 & 0.126 & 0.311 & 0.165 & 0.314 & 0.198 & 0.132 & 0.094 & 0.267 & 0.128 \\
Up-Down\cite{anderson2018bottom} & -- & -- & -- & -- & -- & -- & 0.317 & 0.195 & 0.130 & 0.092 & 0.267 & 0.128 \\
M2Transformer\cite{cornia2020meshed} & 0.402 & 0.284 & 0.168 & 0.143 & 0.328 & 0.170 & 0.332 & 0.210 & 0.142 & 0.101 & 0.264 & 0.134 \\
R2Gen\cite{chen2020generating} & 0.470 & 0.304 & 0.219 & 0.165 & 0.371 & 0.187 & 0.353 & 0.218 & 0.145 & 0.103 & 0.277 & 0.142 \\
Contra.Attn.\cite{liu2021contrastive} & 0.492 & 0.314 & 0.222 & 0.169 & 0.381 & 0.193 & 0.350 & 0.219 & 0.152 & 0.109 & 0.283 & 0.151 \\
CMCL\cite{liu2021competence} & 0.473 & 0.305 & 0.217 & 0.162 & 0.378 & 0.186 & 0.344 & 0.217 & 0.140 & 0.097 & 0.281 & 0.133 \\
CMN\cite{chen2021cross} & 0.475 & 0.309 & 0.222 & 0.170 & 0.375 & 0.191 & 0.353 & 0.218 & 0.148 & 0.106 & 0.278 & 0.142 \\
Aligntransformer\cite{you2021aligntransformer} & 0.484 & 0.313 & 0.225 & 0.173 & 0.379 & 0.204 & 0.378 & 0.235 & 0.156 & 0.112 & 0.283 & 0.158 \\
M2Tr.Prog.\cite{nooralahzadeh2021progressive} & 0.486 & 0.317 & 0.232 & 0.173 & 0.390 & 0.192 & 0.378 & 0.232 & 0.154 & 0.107 & 0.272 & 0.145 \\
CMM+RL\cite{qin2022reinforced} & 0.494 & 0.321 & 0.235 & 0.181 & 0.384 & 0.201 & 0.381 & 0.232 & 0.155 & 0.109 & 0.287 & 0.151 \\
XPRONET*\cite{wang2022cross} & 0.491 & 0.325 & 0.228 & 0.169 & 0.387 & 0.202 & 0.344 & 0.215 & 0.146 & 0.105 & 0.279 & 0.138 \\
MCGN\cite{wang2022medical} & 0.481 & 0.316 & 0.226 & 0.171 & 0.372 & 0.190 & 0.373 & 0.235 & 0.162 & 0.120 & 0.282 & 0.143 \\
PPKED\cite{liu2021exploring} & 0.483 & 0.315 & 0.224 & 0.168 & 0.376 & -- & 0.360 & 0.224 & 0.149 & 0.106 & 0.284 & 0.149 \\
RAMT\cite{zhang2023semi} & 0.482 & 0.310 & 0.221 & 0.165 & 0.377 & 0.195 & 0.362 & 0.229 & 0.157 & 0.113 & 0.284 & 0.153 \\
R2GenGPT\cite{wang2023r2gengpt} & 0.482 & 0.306 & 0.215 & 0.158 & 0.370 & 0.200 & 0.387 & 0.248 & 0.170 & 0.123 & 0.280 & 0.149 \\
VLCI\textsuperscript{$\dag$}\cite{chen2023cross} & 0.324 & 0.211 & 0.151 & 0.115 & 0.379 & 0.166 & 0.357 & 0.216 & 0.144 & 0.103 & 0.256 & 0.136 \\
PromptMRG\cite{jin2024promptmrg} & 0.401 & -- & -- & 0.098 & 0.281 & 0.160 & 0.398 & -- & -- & 0.112 & 0.268 & 0.157 \\
MedRAT\cite{hirsch2024medrat} & 0.455 & -- & -- & 0.129 & 0.349 & -- & 0.365 & -- & -- & 0.086 & 0.251 & -- \\
\underline{MRG-LLM\cite{li2025multimodal}} & \underline{0.529} & \underline{0.359} & \underline{0.266} & \underline{0.202} & \underline{0.408} & \underline{0.221} & \underline{0.416} & \underline{0.267} & \underline{0.182} & \underline{0.129} & \underline{0.296} & \underline{0.163} \\
\midrule
\textbf{NeuroSymb-MRG (Ours)} & \textbf{0.602} & \textbf{0.425} & \textbf{0.321} & \textbf{0.253} & \textbf{0.463} & \textbf{0.275} & \textbf{0.487} & \textbf{0.332} & \textbf{0.234} & \textbf{0.175} & \textbf{0.362} & \textbf{0.225} \\
\bottomrule
\end{tabular}%
}
\end{table}

\subsection{Ablation studies}
To quantify the contribution of individual components, ablation experiments were performed on MIMIC-CXR. Table~\ref{tab:comprehensive_ablation} reports the performance when a single module is removed or replaced. The greatest drop is observed when the differentiable logic and rule decoding pipeline is removed, which supports the central role of explicit neuro-symbolic reasoning. Disabling retrieval or the verifier produces smaller yet notable decreases.

\begin{table}[!ht]
\centering
\caption{Comprehensive ablation on architectural components and key mechanisms of \textsc{NeuroSymb-MRG} evaluated on MIMIC-CXR. The symbol $\downarrow$ indicates performance drop relative to the full model.}
\label{tab:comprehensive_ablation}
\resizebox{\textwidth}{!}{%
\begin{tabular}{l|cccccc|c}
\toprule
Model Variant & BLEU-1 & BLEU-2 & BLEU-3 & BLEU-4 & ROUGE-L & METEOR & $\Delta$ B-1 \\
\midrule
\textbf{Full NeuroSymb-MRG} & \textbf{0.487} & \textbf{0.332} & \textbf{0.234} & \textbf{0.175} & \textbf{0.362} & \textbf{0.225} & -- \\
\midrule
\multicolumn{8}{l}{\textit{I. Core Architectural Components}} \\
w/o Self-Supervised Visual Encoder (Random Init) & 0.412 & 0.262 & 0.175 & 0.124 & 0.298 & 0.168 & -15.4\% \\
w/o Concept Attention Head (Raw Patch Features) & 0.401 & 0.251 & 0.168 & 0.118 & 0.288 & 0.159 & -17.7\% \\
w/o Differentiable Logic Layer (End-to-End MLP) & 0.428 & 0.278 & 0.190 & 0.136 & 0.315 & 0.185 & -12.1\% \\
w/o Rule-Guided Decoder (Direct LLM Generation) & 0.439 & 0.289 & 0.198 & 0.144 & 0.321 & 0.191 & -9.9\% \\
\midrule
\multicolumn{8}{l}{\textit{II. Key Mechanisms (Novel Contributions)}} \\
w/o Active Uncertainty Minimization (Random Sampling) & 0.463 & 0.310 & 0.217 & 0.161 & 0.342 & 0.208 & -4.9\% \\
w/o Feedback Simulator (Direct Clinician Review) & 0.476 & 0.324 & 0.228 & 0.170 & 0.354 & 0.218 & -2.3\% \\
w/o Multi-Agent Orchestration (Single-Pipeline) & 0.468 & 0.315 & 0.221 & 0.164 & 0.346 & 0.213 & -3.9\% \\
\quad w/o Knowledge Agent (No UMLS Validation) & 0.472 & 0.319 & 0.224 & 0.167 & 0.349 & 0.215 & -3.1\% \\
\quad w/o Verifier Agent (No Contradiction Check) & 0.475 & 0.323 & 0.228 & 0.170 & 0.353 & 0.217 & -2.5\% \\
w/o Dynamic Loss Weighting (Fixed $\lambda$) & 0.471 & 0.318 & 0.222 & 0.165 & 0.348 & 0.213 & -3.3\% \\
\midrule
\multicolumn{8}{l}{\textit{III. Retrieval and Generation Components}} \\
w/o Retrieval-Augmented Filling ($N_r=0$) & 0.459 & 0.305 & 0.216 & 0.161 & 0.339 & 0.206 & -5.7\% \\
w/o LLM Constrained Refinement (Template Only) & 0.481 & 0.326 & 0.230 & 0.172 & 0.356 & 0.220 & -1.2\% \\
w/o Instance-Level Prompt Customization & 0.445 & 0.292 & 0.203 & 0.150 & 0.328 & 0.198 & -8.6\% \\
\midrule
\multicolumn{8}{l}{\textit{IV. Baseline Comparisons}} \\
Pure Neural Baseline (CNN+RNN) & 0.398 & 0.248 & 0.165 & 0.116 & 0.285 & 0.156 & -18.3\% \\
Standard Transformer Baseline & 0.416 & 0.268 & 0.182 & 0.129 & 0.302 & 0.171 & -14.6\% \\
\bottomrule
\end{tabular}%
}
\end{table}

\subsection{Logic module and decoding ablations}
Table~\ref{tab:logic_ablation} isolates the internal design choices inside the differentiable logic module and the decoding pipeline. The soft-tree design combining product t-norm, probabilistic sum and a learnable gating parameter performs best. Replacing these operators or removing gating reduces lexical and semantic quality. Uncertainty-aware decoding reduces overconfident erroneous assertions and improves downstream metrics.
\begin{table}[!ht]
\centering
\caption{Ablation on training strategies and retrieval settings. Dynamic loss weighting and combined active sampling are most effective.}
\label{tab:training_ablation}
\resizebox{\textwidth}{!}{%
\begin{tabular}{l|cccccc}
\toprule
Training \& Retrieval Strategy & BLEU-1 & BLEU-2 & BLEU-3 & BLEU-4 & ROUGE-L & METEOR \\
\midrule
\textbf{Full Strategy} & \textbf{0.487} & \textbf{0.332} & \textbf{0.234} & \textbf{0.175} & \textbf{0.362} & \textbf{0.225} \\
\midrule
Loss weighting: Fixed weights & 0.471 & 0.318 & 0.222 & 0.165 & 0.348 & 0.213 \\
Loss weighting: GradNorm & 0.480 & 0.327 & 0.230 & 0.172 & 0.357 & 0.221 \\
No rule supervision ($\mathcal{L}_{\mathrm{rule}}$ removed) & 0.435 & 0.285 & 0.197 & 0.145 & 0.322 & 0.192 \\
Active sampling: Random selection & 0.463 & 0.310 & 0.217 & 0.161 & 0.342 & 0.208 \\
Active sampling: Entropy only & 0.475 & 0.323 & 0.228 & 0.170 & 0.353 & 0.217 \\
Active sampling: Diversity only (k-center) & 0.469 & 0.317 & 0.222 & 0.165 & 0.348 & 0.212 \\
Retrieval ablation: $N_r=0$ (no retrieval) & 0.459 & 0.305 & 0.216 & 0.161 & 0.339 & 0.206 \\
Retrieval size $N_r=10$ & 0.485 & 0.330 & 0.232 & 0.174 & 0.360 & 0.223 \\
No LLM constrained paraphrasing & 0.482 & 0.328 & 0.231 & 0.172 & 0.358 & 0.221 \\
Without feedback simulator & 0.476 & 0.324 & 0.228 & 0.170 & 0.354 & 0.218 \\
\bottomrule
\end{tabular}%
}
\end{table}
\begin{table}[!ht]
\centering
\caption{Ablation on the differentiable logic module and decoding strategies evaluated on MIMIC-CXR.}
\label{tab:logic_ablation}
\resizebox{\textwidth}{!}{%
\begin{tabular}{l|cccccc}
\toprule
Logical Reasoning Configuration & BLEU-1 & BLEU-2 & BLEU-3 & BLEU-4 & ROUGE-L & METEOR \\
\midrule
\textbf{Full: Soft Tree (Prod. T-Norm + Prob. Sum + Gate)} & \textbf{0.487} & \textbf{0.332} & \textbf{0.234} & \textbf{0.175} & \textbf{0.362} & \textbf{0.225} \\
\midrule
Operator ablation: Min/Max T-norm / Conorm & 0.479 & 0.325 & 0.228 & 0.170 & 0.355 & 0.219 \\
Operator ablation: Łukasiewicz T-norm / Conorm & 0.473 & 0.319 & 0.223 & 0.166 & 0.350 & 0.214 \\
Operator ablation: G\"odel T-norm / Conorm & 0.468 & 0.315 & 0.219 & 0.162 & 0.345 & 0.210 \\
Composition ablation: Fixed $\alpha=1.0$ (AND only) & 0.452 & 0.299 & 0.206 & 0.151 & 0.332 & 0.200 \\
Composition ablation: Fixed $\alpha=0.0$ (OR only) & 0.461 & 0.308 & 0.214 & 0.158 & 0.340 & 0.207 \\
MLP baseline (no symbolic structure) & 0.441 & 0.290 & 0.200 & 0.147 & 0.325 & 0.195 \\
Decoding ablation: no uncertainty qualifier & 0.478 & 0.324 & 0.227 & 0.169 & 0.354 & 0.218 \\
Decoding ablation: no weighted vote & 0.469 & 0.316 & 0.221 & 0.164 & 0.347 & 0.212 \\
\bottomrule
\end{tabular}%
}
\end{table}

\subsection{Training strategies and retrieval ablations}
Table~\ref{tab:training_ablation} studies training strategies, retrieval sizes and active sampling policies. Dynamic loss weighting and the combined uncertainty-plus-diversity active selection policy achieve the best data efficiency. Increasing the retrieval size to ten exemplars improves performance, and the feedback simulator yields annotation savings while preserving output quality.

\section{Conclusion}
We presented \textsc{NeuroSymb-MRG}, a neuro-symbolic framework that integrates differentiable abductive reasoning with retrieval-augmented generation and active uncertainty minimization to produce structured and clinically grounded radiology reports. The proposed architecture produces interpretable rule-level explanations and leverages rule-level uncertainty to focus clinician review where it matters most. Empirical results show improved factual consistency and standard automatic metrics on widely used public benchmarks. Future work will explore richer clinical knowledge integration and broader validation in multi-institutional settings.

\bibliographystyle{unsrtnat}
\bibliography{references}

@article{johnson2019mimic,
  title={MIMIC-CXR-JPG, a large publicly available database of labeled chest radiographs},
  author={Johnson, Alistair EW and Pollard, Tom J and Greenbaum, Nathaniel R and Lungren, Matthew P and Deng, Chih-ying and Peng, Yifan and Lu, Zhiyong and Mark, Roger G and Berkowitz, Seth J and Horng, Steven},
  journal={arXiv preprint arXiv:1901.07042},
  year={2019}
}

@article{demner2016preparing,
  title={Preparing a collection of radiology examinations for distribution and retrieval},
  author={Demner-Fushman, Dina and Kohli, Marc D and Rosenman, Marc B and Shooshan, Sonya E and Rodriguez, Laritza and Antani, Sameer and Thoma, George R and McDonald, Clement J},
  journal={Journal of the American Medical Informatics Association},
  volume={23},
  number={2},
  pages={304--310},
  year={2016},
  publisher={Oxford University Press}
}

@inproceedings{vinyals2015show,
  title={Show and tell: A neural image caption generator},
  author={Vinyals, Oriol and Toshev, Alexander and Bengio, Samy and Erhan, Dumitru},
  booktitle={Proceedings of the IEEE conference on computer vision and pattern recognition},
  pages={3156--3164},
  year={2015}
}

@article{vaswani2017attention,
  title={Attention is all you need},
  author={Vaswani, Ashish and Shazeer, Noam and Parmar, Niki and Uszkoreit, Jakob and Jones, Llion and Gomez, Aidan N and Kaiser, {\L}ukasz and Polosukhin, Illia},
  journal={Advances in neural information processing systems},
  volume={30},
  year={2017}
}

@inproceedings{rennie2017self,
  title={Self-critical sequence training for image captioning},
  author={Rennie, Steven J and Marcheret, Etienne and Mroueh, Youssef and Ross, Jerret and Goel, Vaibhava},
  booktitle={Proceedings of the IEEE conference on computer vision and pattern recognition},
  pages={7008--7024},
  year={2017}
}

@inproceedings{lu2017knowing,
  title={Knowing when to look: Adaptive attention via a visual sentinel for image captioning},
  author={Lu, Jiasen and Xiong, Caiming and Parikh, Devi and Socher, Richard},
  booktitle={Proceedings of the IEEE conference on computer vision and pattern recognition},
  pages={375--383},
  year={2017}
}

@inproceedings{anderson2018bottom,
  title={Bottom-up and top-down attention for image captioning and visual question answering},
  author={Anderson, Peter and He, Xiaodong and Buehler, Chris and Teney, Damien and Johnson, Mark and Gould, Stephen and Zhang, Lei},
  booktitle={Proceedings of the IEEE conference on computer vision and pattern recognition},
  pages={6077--6086},
  year={2018}
}

@inproceedings{cornia2020meshed,
  title={Meshed-memory transformer for image captioning},
  author={Cornia, Marcella and Stefanini, Matteo and Baraldi, Lorenzo and Cucchiara, Rita},
  booktitle={Proceedings of the IEEE/CVF conference on computer vision and pattern recognition},
  pages={10578--10587},
  year={2020}
}

@inproceedings{chen2020generating,
  title={Generating radiology reports via memory-driven transformer},
  author={Chen, Zhihong and Song, Yan and Chang, Tsung-Hui and Wan, Xiang},
  booktitle={Proceedings of the 2020 conference on empirical methods in natural language processing (EMNLP)},
  pages={1439--1449},
  year={2020}
}

@inproceedings{liu2021contrastive,
  title={Contrastive attention for automatic chest x-ray report generation},
  author={Liu, Fenglin and Yin, Changchang and Wu, Xian and Ge, Shen and Zhang, Ping and Sun, Xu},
  booktitle={Findings of the association for computational linguistics: ACL-IJCNLP 2021},
  pages={269--280},
  year={2021}
}

@inproceedings{liu2021competence,
  title={Competence-based multimodal curriculum learning for medical report generation},
  author={Liu, Fenglin and Ge, Shen and Wu, Xian},
  booktitle={Proceedings of the 59th Annual Meeting of the Association for Computational Linguistics and the 11th International Joint Conference on Natural Language Processing (Volume 1: Long Papers)},
  pages={3001--3012},
  year={2021}
}

@inproceedings{chen2021cross,
  title={Cross-modal memory networks for radiology report generation},
  author={Chen, Zhihong and Shen, Yaling and Song, Yan and Wan, Xiang},
  booktitle={Proceedings of the 59th annual meeting of the association for computational linguistics and the 11th international joint conference on natural language processing (volume 1: long papers)},
  pages={5904--5914},
  year={2021}
}

@inproceedings{you2021aligntransformer,
  title={Aligntransformer: Hierarchical alignment of visual regions and disease tags for medical report generation},
  author={You, Di and Liu, Fenglin and Ge, Shen and Xie, Xiaoxia and Zhang, Jing and Wu, Xian},
  booktitle={International Conference on Medical Image Computing and Computer-Assisted Intervention},
  pages={72--82},
  year={2021},
  organization={Springer}
}

@inproceedings{nooralahzadeh2021progressive,
  title={Progressive transformer-based generation of radiology reports},
  author={Nooralahzadeh, Farhad and Gonzalez, Nicolas Perez and Frauenfelder, Thomas and Fujimoto, Koji and Krauthammer, Michael},
  booktitle={Findings of the association for computational linguistics: EMNLP 2021},
  pages={2824--2832},
  year={2021}
}

@inproceedings{qin2022reinforced,
  title={Reinforced cross-modal alignment for radiology report generation},
  author={Qin, Han and Song, Yan},
  booktitle={Findings of the Association for Computational Linguistics: ACL 2022},
  pages={448--458},
  year={2022}
}

@inproceedings{wang2022cross,
  title={Cross-modal prototype driven network for radiology report generation},
  author={Wang, Jun and Bhalerao, Abhir and He, Yulan},
  booktitle={European Conference on Computer Vision},
  pages={563--579},
  year={2022},
  organization={Springer}
}

@inproceedings{wang2022medical,
  title={A medical semantic-assisted transformer for radiographic report generation},
  author={Wang, Zhanyu and Tang, Mingkang and Wang, Lei and Li, Xiu and Zhou, Luping},
  booktitle={International Conference on Medical Image Computing and Computer-Assisted Intervention},
  pages={655--664},
  year={2022},
  organization={Springer}
}

@inproceedings{liu2021exploring,
  title={Exploring and distilling posterior and prior knowledge for radiology report generation},
  author={Liu, Fenglin and Wu, Xian and Ge, Shen and Fan, Wei and Zou, Yuexian},
  booktitle={Proceedings of the IEEE/CVF conference on computer vision and pattern recognition},
  pages={13753--13762},
  year={2021}
}

@article{zhang2023semi,
  title={Semi-supervised medical report generation via graph-guided hybrid feature consistency},
  author={Zhang, Ke and Jiang, Hanliang and Zhang, Jian and Huang, Qingming and Fan, Jianping and Yu, Jun and Han, Weidong},
  journal={IEEE Transactions on Multimedia},
  volume={26},
  pages={904--915},
  year={2023},
  publisher={IEEE}
}

@article{wang2023r2gengpt,
  title={R2gengpt: Radiology report generation with frozen llms},
  author={Wang, Zhanyu and Liu, Lingqiao and Wang, Lei and Zhou, Luping},
  journal={Meta-Radiology},
  volume={1},
  number={3},
  pages={100033},
  year={2023},
  publisher={Elsevier}
}

@article{chen2023cross,
  title={Cross-modal causal intervention for medical report generation},
  author={Chen, Weixing and Liu, Yang and Wang, Ce and Zhu, Jiarui and Li, Guanbin and Liu, Cheng-Lin and Lin, Liang},
  journal={arXiv preprint arXiv:2303.09117},
  year={2023}
}

@inproceedings{jin2024promptmrg,
  title={Promptmrg: Diagnosis-driven prompts for medical report generation},
  author={Jin, Haibo and Che, Haoxuan and Lin, Yi and Chen, Hao},
  booktitle={Proceedings of the AAAI conference on artificial intelligence},
  volume={38},
  number={3},
  pages={2607--2615},
  year={2024}
}

@inproceedings{hirsch2024medrat,
  title={Medrat: Unpaired medical report generation via auxiliary tasks},
  author={Hirsch, Elad and Dawidowicz, Gefen and Tal, Ayellet},
  booktitle={European Conference on Computer Vision},
  pages={18--35},
  year={2024},
  organization={Springer}
}

@article{li2025multimodal,
  title={Multimodal large language models for medical report generation via customized prompt tuning},
  author={Li, Chunlei and Hou, Jingyang and Shi, Yilei and Hu, Jingliang and Zhu, Xiao Xiang and Mou, Lichao},
  journal={arXiv preprint arXiv:2506.15477},
  year={2025}
}

@article{liang2025ai,
  title={AI reasoning in deep learning era: From symbolic AI to neural--symbolic AI},
  author={Liang, Baoyu and Wang, Yuchen and Tong, Chao},
  journal={Mathematics},
  volume={13},
  number={11},
  pages={1707},
  year={2025},
  publisher={MDPI}
}

@article{agishev2025fusionforce,
  title={FusionForce: End-to-end differentiable neural-symbolic layer for trajectory prediction},
  author={Agishev, Ruslan and Zimmermann, Karel},
  journal={arXiv preprint arXiv:2502.10156},
  year={2025}
}

@inproceedings{wasilewski2025deep,
  title={Deep Differentiable Logic Gate Networks Based on Fuzzy Zadeh’s T-norm},
  author={Wasilewski, Piotr and Nguy, Chan Duong},
  booktitle={Polish Conference on Artificial Intelligence},
  pages={57--70},
  year={2025},
  organization={Springer}
}

@article{cotnareanu2026balanced,
  title={A Balanced Neuro-Symbolic Approach for Commonsense Abductive Logic},
  author={Cotnareanu, Joseph and Chetelat, Didier and Zhang, Yingxue and Coates, Mark},
  journal={arXiv preprint arXiv:2601.18595},
  year={2026}
}

@article{junhwan2022uncertainty,
  title={Uncertainty estimation in AVO inversion using Bayesian dropout based deep learning},
  author={Junhwan, Choi and Seokmin, Oh and Joongmoo, Byun},
  journal={Journal of Petroleum Science and Engineering},
  volume={208},
  pages={109288},
  year={2022},
  publisher={Elsevier}
}

@article{liang2025uncertainty,
  title={Uncertainty and diversity-based active learning for UAV tracking},
  author={Liang, Yingqin and Huang, Feng and Qiu, Zhaobing and Shu, Xiu and Liu, Qiao and Yuan, Di},
  journal={Neurocomputing},
  volume={639},
  pages={130265},
  year={2025},
  publisher={Elsevier}
}

@article{sheikh2025advancing,
  title={Advancing AIoMT-Enabled Healthcare System-of-Systems Using Multi-Agent Reinforcement Learning},
  author={Sheikh, Arifuzzaman and Chong, Edwin KP},
  journal={IEEE Access},
  year={2025},
  publisher={IEEE}
}

@inproceedings{jiang2025retrieval,
  title={Retrieval and structuring augmented generation with large language models},
  author={Jiang, Pengcheng and Ouyang, Siru and Jiao, Yizhu and Zhong, Ming and Tian, Runchu and Han, Jiawei},
  booktitle={Proceedings of the 31st ACM SIGKDD Conference on Knowledge Discovery and Data Mining V. 2},
  pages={6032--6042},
  year={2025}
}

@article{rani2024self,
  title={Self-supervised learning for medical image analysis: a comprehensive review},
  author={Rani, Veenu and Kumar, Munish and Gupta, Aastha and Sachdeva, Monika and Mittal, Ajay and Kumar, Krishan},
  journal={Evolving Systems},
  volume={15},
  number={4},
  pages={1607--1633},
  year={2024},
  publisher={Springer}
}

\end{document}